\documentclass[a4paper]{report}
\usepackage[utf8]{inputenc}
\usepackage[T1]{fontenc}
\usepackage{RJournal}
\usepackage{amsmath,amssymb,array}
\usepackage{booktabs}

\newtheorem{definition}{Definition}[section]
\usepackage[ruled]{algorithm2e}

\begin{document}


\sectionhead{conformalClassification}
\begin{article}
\title{conformalClassification: A Conformal Prediction  R Package for Classification 
}
\author{by Niharika Gauraha and Ola Spjuth}

\maketitle

\abstract{
The conformalClassification package implements Transductive Conformal Prediction (TCP) and Inductive Conformal Prediction (ICP) for classification problems. Conformal Prediction (CP) is a framework that complements the predictions of machine learning algorithms with reliable measures of confidence. TCP gives results with higher validity than ICP, however ICP is computationally faster than TCP. The package  conformalClassification is built upon the random forest method, where votes of the random forest for each class are considered as the conformity scores for each data point. Although the main aim of the conformalClassification package is to generate CP errors (p-values) for classification problems, the package also implements various diagnostic measures such as deviation from validity, error rate, efficiency, observed fuzziness and calibration plots. In future releases, we plan to extend  the package to use other machine learning algorithms, (e.g. support vector machines) for model fitting.
}

\section{Introduction}

Conformal predictors are confidence predictors that result in prediction sets for all confidence levels. Thus, Conformal Prediction (CP) is a framework that complements the predictions of machine learning algorithms with reliable measures of confidence. 
Transductive Conformal Prediction (TCP) works in an on-line transductive setting, such that learning and prediction occur simultaneously. In this sense confidence in a prediction is tailored both to the previously seen objects (whose features and labels are known) and to the features of the new object, whose label is to be predicted. By conditioning on the new objects conformal predictors take account of how difficult a particular object is to label and adjust their confidence in the prediction accordingly, as opposed to having an overall error rate for labelling all new objects \cite{vovk2005algorithmic}. The output from a TCP algorithm is thus a point prediction and a region prediction, such as a 95\% prediction region which, under minimal assumptions, contains the true label with a probability of at least 0.95 \cite{shafer2008tutorial}. The method for point prediction embedded within the CP framework can be almost any machine learning algorithm, such as random forests, support vector machines or neural networks. Based on the chosen learning algorithm a nonconformity measure is created which evaluates the ``strangeness" of the new object relative to those previously seen. The TCP algorithm utilizes this nonconformity score to define the appropriate prediction region \cite{shafer2008tutorial}.

The fully on-line mode of TCP can be very computationally demanding (with the learning algorithm updated for each new data point). The theory however extends easily to the off-line inductive (batch) mode giving rise to what we refer to here as Inductive Conformal Prediction (ICP). CP has been used in moderately sized problems, e.g. to predict quantitative structure-activity relationships of molecules \cite{norinder2014introducing}, to assess complication risks following coronary procedures \cite{balasubramanian2014conformal} and to detect anomalies in fishing vessel trajectories \cite{smith2015conformal}. It has also been shown to scale up well on a distributed computing implementation to very large datasets, such as the Higgs boson dataset of 11 million data points \cite{capuccini2015conformal}, the largest binary classification dataset in the UCI machine learning repository \cite{bache2013uci}.

In the release version of conformalClassification we use random forests as the underlying machine learning method, where the vote for each class – the ratio between the number of trees in the forest voting for a given class divided by the total number of tree – gives the conformity score for each data point. 

\section{Background and Notations}
This section gives a brief background about CP and fixes notations and definitions used throughout the article.

The object space is denoted by $\mathcal{X} \in \mathbb{R}^p$, where $p$ is the number of features, and  label space is denoted by $\mathcal{Y} \in (1,2,...,l)$, where $l$ is the number of class labels. We assume that each observation consists of an object and its label, and its space is given as $\mathcal{Z} := \mathcal{X} \times \mathcal{Y}$. 
The typical classification problem is, given a training dataset $Z = \{ z_1 , ..., z_n \} $ -- where $n$ is the number of observations in the training set, and each observation $z_i = (x_i, y_i)$ is a labeled observation -- we want to predict the label of a new observation $x_{new}$ whose label is unknown. The exchangeability \citep{shafer2008tutorial} of observations is assumed throughout the paper.

The nonconformity measure is a function that measures the disagreement of possible labels of a test object with respect to an observed distribution. 

\begin{definition} [Nonconformity Measure]
A nonconformity measure is a measurable function $\mathcal{A} : \mathcal{Z} \times \mathcal{Z}  \rightarrow \mathbb{R}$ such that $\mathcal{A}(Z_1 , Z_2 )$ does not depend on the ordering of observations in the set $Z_1$. 
\end{definition}

The nonconformity scores are most often derived from the underlying algorithms used for point prediction.  For classification problems, the error rate may be higher in some classes than others, to overcome this issue the nonconformity scores are applied on per class basis, this is referred to as Mondrian CP \cite{norinder2014introducing}. Alternatively, the conformity measure can be defined as, $1 - \mathcal{A} $(nonconformity measure).

A natural conformity measure for classification problems using random forests method \cite{Breiman} is the proportion of votes for each class, the ratio between the number of trees in the forest voting for a given class divided by the total number of trees.

\begin{align} \label{eq:def_nonconformity}
\begin{split}
\alpha_i(y) &= \frac{\#\text{trees voting for class y}}{\#\text{of trees}}
\end{split}
\end{align}

We denote by $\alpha_i(y)$, the conformity score for $i^{th}$ observation for class $y$. Each component  $\alpha_i(y)$ that corresponds to the sample $(x_i,y_i)$ is computed by
equation (\ref{eq:def_nonconformity}) based on the augmented sample  $\{ z_1 , ..., z_n, z_{n+1}=(x_{new},y) \}$. Then p-value as defined below, \cite{vovk2005algorithmic}, describes the lack of conformity of the  new observation $x_{new}$ to the training set $Z$. 
 \begin{align*}
 p_y &= \frac{| \{ z_i \in Z : y_i=y, \alpha_i(y) < \alpha_{new}(y) \} | + u_i* | \{ z_i \in Z : y_i=y, \alpha_i(y) = \alpha_{new}(y)\} |}{n_y+1} \\
 \end{align*}
 where $u_i \sim U[0,1]$, $n_y$ denotes the number of observations having the true label as class-y in the training set. The p-value $p(y)=p_y, \ y \in Y$ lies in $ \left( \frac{1}{n_y+1},1 \right)$. The smaller the $p(y)$  is, the less likely the true pair is $(x_{new},y)$. Multiplying the borderline cases by $u_i$ results in what are known as smoothed conformal predictors \citep{vovk2005algorithmic}.
 
\begin{definition}[Transductive Conformity Prediction (TCP)]
Given a training dataset $Z$ and a new observation $x_{new}$, the transductive conformal predictor (TCP ), corresponding to a nonconformity measure $\mathcal{A}$, checks each of a set of hypothesis (for all possible labels) for the new observation $x_{new}$, assigns to it a p-value,  and finds the prediction region for the test set $x_{new}$ at a significance level $\epsilon \in (0, 1)$.
\end{definition}

The predicted region of a test observation is a subset of $\mathcal{Y}$ , denoted as $\Gamma^{\epsilon} = \{ y \mid p_y > \epsilon \}$, at a significance level $\epsilon \in (0, 1)$. A prediction region  $\Gamma^{\epsilon} = \{ y \mid p_y > \epsilon \}$ contains the
true value of a test observation with probability at least $1 -\epsilon$. The prediction region $\Gamma^{\epsilon}$ can be any one of the following:
\begin{enumerate}
\item Empty, when $|\Gamma^{\epsilon}| = 0$.
\item Singleton, when $|\Gamma^{\epsilon}| = 1$.
\item Multiple, when $|\Gamma^{\epsilon}| >1$.
\end{enumerate} 

\begin{algorithm}[H]
 \caption{\textbf{TCP}} \label{algo:TCP}
 \textbf{Input:}{ (training dataset:$Z$, test data:$x_{new}$, label set:$Y$, a nonconformity measure:$\mathcal{A})$}\\
 \textbf{Output:}{\textbf{ p-values} }\\
 \For{each $y \in \mathcal{Y}$ }{
 	$z_{n+1} = (x_{new},y) $;\\
 	$Z^* = (Z,z_{n+1})$ ;\\
 	Compute the transductive nonconformity scores:\\
 	 $\alpha_i = \mathcal{A}(Z^*, z_i)$ for each $z_i \in Z^*$;\\
 	 \textbf{\\}
	Compute p-value: $ p(y) = \frac{| \{ i \in \{1,..,n+1\} : y_i=y, \alpha_i(y) < \alpha_{new}(y) \} | + u_i*| \{ i \in \{1,..,n+1\} : y_i=y, \alpha_i(y) = \alpha_{new}(y) \} |}{n_y + 1}$;\\
  }

 $\textbf{p-values} = \{ p(y)| y \in \mathcal{Y}\}$;\\
 \textbf{return \textbf{p-values}};\\
 \end{algorithm}

For further details on TCP, we refer to \cite{vapnik1998statistical}, \cite{shafer2008tutorial}, \cite{vovk2005algorithmic} and \cite{balasubramanian2014conformal}.

The computational expense of TCP, whereby the prediction rule is updated for each new example for each class label, may be computationally intractable for large datasets. To address this issue the batch-mode ICP method was introduced. For ICP, the training set $Z$ is partitioned into two different sets: the proper training set, ${Z_p = z 1 , . . . , z_q }$ of size $q$, and the calibration set ${Z_c = z_{q+1} , . . . , z_n }$ of size $n-q$. ICP relies on the idea that how well the calibration set conforms to the proper training set. The ICP p-value is then computed as 
 \begin{align*}
 p_y &= \frac{| \{ z_i \in Z_c : y_i=y, \alpha_i(y) < \alpha_{new}(y) \} | + u_i* | \{ z_i \in Z_c : y_i=y, \alpha_i(y) = \alpha_{new}(y)\} |}{n_y+1}, \\
 \end{align*}
where $n_y$ denotes the number of observations having the true label as class-y in the calibration set.

\begin{algorithm}[H]
 \caption{\textbf{ICP}} \label{algo:ICP}
 \textbf{Input:}{ (training dataset:$Z$, test data:$x_{new}$, label set:$Y$, a nonconformity measure:$\mathcal{A})$}\\
 \textbf{Output:}{\textbf{ p-values} }\\
 partition $Z$ into proper training set $Z_p$ and calibration set $Z_c$ \\
 Compute nonconformity scores:\\
 	 $\alpha_i = \mathcal{A}(Z_p, z_i)$ for each $z_i \in Z_c$;\\
 	 \textbf{\\}
 	 Compute nonconformity score for test observation:	
 	 $\alpha_{new} = \mathcal{A}(Z_p, (x_{new}, y) )$ for each $y \in Y$ \\
	Compute p-values: \\
	$ p(y) = \frac{| \{ z_i \in Z_c : y_i=y, \alpha_i(y) < \alpha_{new}(y) \} | + u_i* | \{ z_i \in Z_c : y_i=y, \alpha_i(y) = \alpha_{new}(y)\} |}{n_y+1}$;\\
   
 $\textbf{p-values} = \{ p(y)| y \in Y \}$;\\
 \textbf{return \textbf{p-values}};\\
 \end{algorithm}
 
To evaluate the performance of conformal predictors, we consider the following criterion: error rate, validity, efficiency and observed fuzziness. 
A predictor makes an error when the predicted region does not contain the true label, that is $ y \not\in |\Gamma^{\epsilon}|$. Given a training dataset $Z$ and an external test set $Z_T$,  and $|Z_T| = m$. Suppose that a conformal predictor gives prediction regions as $\Gamma_1^{\epsilon}, ...., \Gamma_m^{\epsilon}$, then the error rate is defined as follows.

\begin{definition}[Error rate]
\begin{align} \label{eq:errorRate}
		ER^{\epsilon} &= \frac{ 1}{m} \sum\limits_{i=1}^{m} \textbf{I}_{ \{y_i \not\in \Gamma_i^{\epsilon} \} },		
\end{align}	
where $y_i$ is the true class label of the $i^{th}$ test case and $\textbf{I}$ is an indicator function. 	
\end{definition}
The efficiency can be computed as the ratio of predictions with more than one class over number of observations in the test set.
\begin{definition}[Efficiency]
\begin{align} \label{eq:efficiency}
		EFF^{\epsilon}  = \frac{ 1}{m} { \sum\limits_{i=1}^{k} I_{(|\Gamma^{\epsilon}| >1 )}}
\end{align}	 
\end{definition}

The deviation from exact validity can be computed as (\cite{carlsson2017comparing}) the Euclidean norm of the difference of the observed error and the expected error for a given set of predefined significance levels. Let us assume a set of significance levels $\epsilon = \{ \epsilon_1, ..., \epsilon_k \}$, then the formula for the validity can be given as follows.

\begin{definition}[Deviation from Validity]
\begin{align} \label{eq:validity}
		VAL = \sqrt{ \sum\limits_{i=1}^{k} (ER^{\epsilon_i} -\epsilon_i)^2 }
\end{align}	 
\end{definition}
The Observed fuzziness is defined as the sum of all p-values for the incorrect class labels. 

\begin{definition}[Observed Fuzziness]
\begin{align} \label{eq:ObsFuzz}
	ObsFuzz =\frac{ 1}{m} \sum\limits_{i=1}^{m} \sum\limits_{y_i \neq y }  p_i^y,		
\end{align}
\end{definition}
We note that for the above measure of performances, smaller values are preferable. 







\section{Conclusions}

The conformalClassification package implements Transductive Conformal Prediction and Inductive Conformal Prediction for Classification problems using Random Forests as the underlying machine learning algorithm. 

\section{Future Development}

In future releases, we plan to extend package to use other machine learning algorithms, (e.g.  support vector machines) for model fitting.

\section{Acknowledgements}
The authors acknowledge UPPMAX, Uppsala Multidisciplinary Centre for Advanced Computational Science for providing computational resources. The authors would also like to thank Philip J. Harrison for comments and recommendations during the preparation of this manuscript and R package.
\bibliographystyle{plain}
\bibliography{RJreferences}

\begin{thebibliography}{10}
\providecommand{\natexlab}[1]{#1}
\providecommand{\url}[1]{\texttt{#1}}
\expandafter\ifx\csname urlstyle\endcsname\relax
  \providecommand{\doi}[1]{doi: #1}\else
  \providecommand{\doi}{doi: \begingroup \urlstyle{rm}\Url}\fi

\bibitem[Bache and Lichman(2013)]{bache2013uci}
K.~Bache and M.~Lichman.
\newblock Uci machine learning repository.
\newblock 2013.

\bibitem[Balasubramanian et~al.(2014)Balasubramanian, Ho, and
  Vovk]{balasubramanian2014conformal}
V.~Balasubramanian, S.-S. Ho, and V.~Vovk.
\newblock \emph{Conformal Prediction for Reliable Machine Learning: Theory,
  Adaptations and Applications}.
\newblock Newnes, 2014.

\bibitem[Breiman(2001)]{Breiman}
L.~Breiman.
\newblock Random forests.
\newblock \emph{Mach. Learn.}, 45\penalty0 (1):\penalty0 5--32, Oct. 2001.
\newblock ISSN 0885-6125.
\newblock \doi{10.1023/A:1010933404324}.
\newblock URL \url{https://doi.org/10.1023/A:1010933404324}.

\bibitem[Capuccini et~al.(2015)Capuccini, Carlsson, Norinder, and
  Spjuth]{capuccini2015conformal}
M.~Capuccini, L.~Carlsson, U.~Norinder, and O.~Spjuth.
\newblock Conformal prediction in spark: Large-scale machine learning with
  confidence.
\newblock In \emph{Big Data Computing (BDC), 2015 IEEE/ACM 2nd International
  Symposium on}, pages 61--67. IEEE, 2015.

\bibitem[Carlsson et~al.(2017)Carlsson, Bendtsen, and
  Ahlberg]{carlsson2017comparing}
L.~Carlsson, C.~Bendtsen, and E.~Ahlberg.
\newblock Comparing performance of different inductive and transductive
  conformal predictors relevant to drug discovery.
\newblock In \emph{Conformal and Probabilistic Prediction and Applications},
  pages 201--212, 2017.

\bibitem[Norinder et~al.(2014)Norinder, Carlsson, Boyer, and
  Eklund]{norinder2014introducing}
U.~Norinder, L.~Carlsson, S.~Boyer, and M.~Eklund.
\newblock Introducing conformal prediction in predictive modeling. a
  transparent and flexible alternative to applicability domain determination.
\newblock \emph{Journal of chemical information and modeling}, 54\penalty0
  (6):\penalty0 1596--1603, 2014.

\bibitem[Shafer and Vovk(2008)]{shafer2008tutorial}
G.~Shafer and V.~Vovk.
\newblock A tutorial on conformal prediction.
\newblock \emph{Journal of Machine Learning Research}, 9\penalty0
  (Mar):\penalty0 371--421, 2008.

\bibitem[Smith et~al.(2015)Smith, Nouretdinov, Craddock, Offer, and
  Gammerman]{smith2015conformal}
J.~Smith, I.~Nouretdinov, R.~Craddock, C.~Offer, and A.~Gammerman.
\newblock Conformal anomaly detection of trajectories with a multi-class
  hierarchy.
\newblock In \emph{International Symposium on Statistical Learning and Data
  Sciences}, pages 281--290. Springer, 2015.

\bibitem[Vapnik and Vapnik(1998)]{vapnik1998statistical}
V.~N. Vapnik and V.~Vapnik.
\newblock \emph{Statistical learning theory}, volume~1.
\newblock Wiley New York, 1998.

\bibitem[Vovk et~al.(2005)Vovk, Gammerman, and Shafer]{vovk2005algorithmic}
V.~Vovk, A.~Gammerman, and G.~Shafer.
\newblock \emph{Algorithmic learning in a random world}.
\newblock Springer Science \& Business Media, 2005.

\end{thebibliography}

\address{Niharika Gauraha\\
  Uppsala University\\
  Uppsala\\
  Sweden\\
  \email{niharika.gauraha@farmbio.uu.se}}

\address{Ola Spjuth\\
  Uppsala University\\
  Uppsala\\
  Sweden\\
  \email{ola.spjuth@farmbio.uu.se}}

\end{article}

\end{document}